\documentclass[format=acmsmall, review=false, screen=true]{acmart}

\usepackage{booktabs} 

\usepackage[ruled]{algorithm2e} 

\SetAlFnt{\small}
\SetAlCapFnt{\small}
\SetAlCapNameFnt{\small}
\SetAlCapHSkip{0pt}
\IncMargin{-\parindent}

\acmDOI{0000001.0000001}

\usepackage{tcolorbox}
\usepackage[utf8]{inputenc}
\usepackage{graphicx}
\usepackage{mathrsfs}
\usepackage{listings}
\title[Open the Black Box]{Open the Black Box\\ Data-Driven Explanation of Black Box Decision Systems}

\author{Dino Pedreschi}
\affiliation{%
  \institution{University of Pisa, Italy}
}
\email{dino.pedreschi@unipi.it}

\author{Fosca Giannotti}
\affiliation{%
  \institution{ISTI-CNR of Pisa, Italy}
}
\email{fosca.giannotti@isti.cnr.it}

\author{Riccardo Guidotti}
\affiliation{%
  \institution{ISTI-CNR \& University of Pisa, Italy}
}
\email{riccardo.guidotti@isti.cnr.it}

\author{Anna Monreale}
\affiliation{%
  \institution{University of Pisa, Italy}
}
\email{anna.monreale@unipi.it}

\author{Luca Pappalardo }
\affiliation{%
  \institution{ISTI-CNR of Pisa, Italy}
}
\email{luca.pappalardo@unipi.it}

\author{Salvatore Ruggieri }
\affiliation{%
  \institution{University of Pisa, Italy}
}
\email{salvatore.ruggieri@unipi.it}

\author{Franco Turini}
\affiliation{%
  \institution{University of Pisa, Italy}
}
\email{franco.turini@unipi.it}

\renewcommand{\shortauthors}{D. Pedreschi et. al.}

\begin{document}

\maketitle

\begin{abstract}
    Black box systems for automated decision making, often based on machine learning over (big) data, map a user’s features into a class or a score without exposing the reasons why. This is problematic not only for lack of transparency, but also for possible biases hidden in the algorithms, due to human prejudices and collection artifacts hidden in the training data, which may lead to unfair or wrong decisions. 
    We introduce the local-to-global framework for black box explanation, a novel approach with promising early results, which paves the road for a wide spectrum of future developments along three dimensions: ($i$) the language for expressing explanations in terms of highly expressive logic-based rules, with a statistical and causal interpretation; ($ii$) the inference of local explanations aimed at revealing the logic of the decision adopted for a specific instance by querying and auditing the black box in the vicinity of the target instance; ($iii$), the bottom-up generalization of the many local explanations into simple global ones, with algorithms that optimize the quality and comprehensibility of explanations. 
\end{abstract}


\section{Introduction}
\label{sec:intro}
The last decade has witnessed the rise of a black box society \cite{pasquale2015black}. Ubiquitous obscure algorithms, often based on sophisticated machine learning models trained on (big) data, which predict behavioural traits of individuals, such as credit risk, health status, personality profile. Black boxes map user features into a class or a score without explaining why, because the decision model is either not comprehensible to stakeholders, or secret. This is worrying not only in terms of the lack of transparency, but also due to the possible biases hidden in the algorithms. Machine learning (ML) constructs predictive models and decision-making systems based on (possibly big) data, i.e., the digital traces of human activities (opinions, movements, lifestyles, etc.). Consequently, these models may reflect human biases and prejudices, as well as collection artifacts, possibly leading to unfair or simply wrong decisions. Many controversial cases have already highlighted that delegating decision-making to black box algorithms is critical in many sensitive domains, including crime prediction, personality scoring, image classification, personal assistance, and more (see box “The danger of black boxes".)

The EU General Data Protection Regulation (GDPR), entered into force in Europe on 25 May 2018, introduces a right of explanation for individuals to obtain “meaningful information of the logic involved” when automated decision making takes place with “legal effects” on individuals “or similarly significantly affect[ing]” them\footnote{\url{http://ec.europa.eu/justice/data-protection/}}. Without an enabling technology capable of explaining the logic of black boxes, this right will either remain “dead letter”, or will just outlaw many such systems \cite{goodman2016eu,comande2017right,wachter2017right}.

Through sophisticated machine learning models trained on massive datasets, we risk creating and using automated decision systems that we do not really understand. This impacts not only information ethics, but also accountability, safety and industrial liability \cite{DBLP:journals/expert/DanksL17,DBLP:conf/sgai/Kingston16,Kroll2017}. Companies increasingly market services and products by embedding machine learning components, often in safety-critical industries such as self-driving cars, robotic assistants, domotic IoT systems, and personalized medicine. Another inherent risk of these components is the possibility of inadvertently making wrong decisions, learned from artifacts or spurious correlations in the training data, such as recognizing an object in a picture by the properties of the background or lighting, due to a systematic bias in training data collection. How can companies trust their products without understanding and validating the underlying rationale of their machine learning components? An explanation technology would be of immense help to companies for creating safer, more trustable products, and better managing any possible liability they may have. Likewise, the use of machine learning models in scientific research, for example in medicine, biology, socio-economic sciences, requires an explanation not only for trust and acceptance of results, but also for the very sake of the openness of scientific discovery and the progress of research. Explanation is at the heart of a responsible, open data science, across multiple industry sectors and scientific disciplines. 

Despite the soaring recent body of research on interpretable ML and explainable AI, a practical, widely applicable technology for black box explanation has not emerged yet. The challenge is hard, as explanations should be sound and complete in statistical and causal terms, and yet comprehensible to multiple stakeholders such as the users subject to the decisions, the developers of the automated decision system, researchers, data scientists and policy makers, authorities and auditors, including regulation and competition commissions, civil rights societies, etc. Stakeholders should be empowered to reason on explanations, to understand how the automated decision-making system works on the basis of the inputs provided by the user; what are the most critical features; whether the system adopts latent features; how a specific decision is taken and on the basis of what rationale/reasons; how the user could get a better decision in the future.

After a succinct, high-level perspective on the booming field of explainable, interpretable machine learning, we focus on the open challenge of how to construct meaningful explanations of black boxes, and delineate a novel research direction suggested by a few recent methods for local explanations, i.e., methods to explain why a certain specific case has received its own classification outcome. 
Starting from these methods, including our own, we propose a new, local-first explanation framework: expressive logic rule languages for inferring local explanations, together with bottom-up generalization methods to aggregate an exhaustive collection of local explanations into a global one, which optimizes jointly both simplicity and fidelity in mimicking a black box. We argue that the local-first approach has the potential to advance the state of art significantly, opening the door to a wide variety of alternative technical solutions.

\begin{tcolorbox}
\paragraph{\textbf{The danger of black boxes.}} Delegating decisions to black boxes may be critical, as the following cases illustrate.

\begin{itemize}
    \item The COMPAS score is a predictive model for the “risk of crime recidivism”, a proprietary secret of Northpointe, Inc. Journalists at propublica.org have shown that the model has a strong ethnic bias: blacks who did not reoffend were classified as high risk twice as much as whites who did not reoffend\footnote{\url{www.propublica.org/article/machine-bias-risk-assessments-in-criminal-sentencing}}.

    \item Three major US credit bureaus, Experian, TransUnion, and Equifax, providing credit scoring for millions of individuals, are often discordant. In a study of 500,000 records, 29\% of consumers received credit scores that differed by at least fifty points between credit bureaus, a difference that may mean tens of thousands of dollars over the life of a mortgage \cite{carter2006credit}. So much variability suggests that the three scoring systems either have a different (undisclosed) bias or are highly arbitrary.

    \item During the 1970s and 1980s, St. George’s Hospital Medical School in London used a computer program for screening job applicants, based on information from applicants’ forms. The program was found to unfairly discriminate against women and ethnic minorities, inferred from surnames and place of birth, lowering their chances of being selected for interview \cite{lowry1988blot}. This shows that automated discrimination is not new and is not necessarily due to machine learning. 

    \item In \cite{ribeiro2016should} the authors show how an accurate but untrustworthy classifier may result from an accidental artifact in the training data. In a task that involved discriminating between wolves and husky dogs in a dataset of images, the resulting deep learning model is shown to classify a wolf based solely on the presence of snow in the background of the picture.
    
    \item A study at Princeton \cite{caliskan2016semantics} shows how text/web corpora contain human biases, such as names that were associated with whites were found to be significantly more associated with pleasant than unpleasant terms, compared to names associated with black people. Therefore, models trained on text data for sentiment or opinion mining have a strong chance of inheriting the prejudices reflected in the data. 

    \item In 2016, Amazon.com used software to determine the areas of the US which it would offer free same-day delivery to\footnote{\url{http://www.businessinsider.com/how-algorithms-can-be-racist-2016-4}}. It turned out that the software inadvertently prevented minority neighbourhoods from participating in the program, often when every surrounding neighbourhood was allowed.
\end{itemize}

Learning from historical data recording human decision making may lead to the discovery of prejudices that are endemic in reality \cite{Barocas2016,pedreshi2008discrimination}, and to assign to such practices the status of general rules, maybe unconsciously, as these rules can be deeply hidden within the learned classifier. Today, we are warned about a rising ``black box society'' \cite{pasquale2015black}, governed by ``secret algorithms protected by industrial secrecy, legal protections, obfuscation, so that intentional or unintentional discrimination becomes invisible and mitigation becomes impossible''.

\end{tcolorbox}

\newpage 

\section{The black-box explanation problem}
\label{sec:exp}

Two different flavors of black-box explanation exist:
\begin{itemize}
    \item the \emph{eXplanation by Design} (XbD) problem: given a dataset of training decision records, how to develop a machine learning decision model \emph{together with} its explanation;
    \item the \emph{Black Box eXplanation} (BBX) problem: given the decision records produced by an inscrutable black box decision model, how to reconstruct an explanation for it.
\end{itemize}
In the XbD problem setting, the data scientist in charge of developing a decision model with machine learning is also supposed to provide an explanation of the model’s logic, in order to prevent the model from making unfair, inaccurate or simply wrong decisions (such as the wolves on the snow) learned from artifacts and biases hidden in the training data and/or amplified or introduced by the learning algorithm. In this scenario, where the data scientist has full control over the model’s creation process, the development of an explanation is essentially a further validation step in assessing the quality of the output model (in addition to testing for accuracy, absence of overfitting, etc.). At the same time, the explanation is an extra deliverable of the learning process, sustaining transparency and the trust of the stakeholders who will be adopting the model.

In the harder BBX problem setting, the data scientist aims to find an explanation for a black box designed by others. In this case, the original dataset on which the black box was trained is generally not known, and neither are the internals of the model. In fact, only the decision behaviour of the black box can, to some extent, be observed. This task has important variants, making the problem of revealing explanations increasingly difficult: can the black box be queried at will to obtain new decision examples, or only a given sample dataset of decision records is available? Is the complete set of features used by the decision model known, or only some of these features?

Although attempts to tackle these problems by means of \emph{interpretable machine learning} and \emph{discrimination-aware data mining} exist for several years now, there has been an exceptional growth of research efforts in the last couple of years, with new emerging keywords such as \emph{black box explanation} and \emph{explainable AI}. We provide a comprehensive, up-to-date survey \cite{guidotti2018survey}, and account here for the major recent trends. Many approaches to the XbD problem attempt at explaining the \textit{global} logic of a black box by an associated interpretable classifier that mimics the black box. These methods are mostly designed for specific machine learning models, i.e., they are not agnostic, and often the interpretable classifier consists in a decision tree or in a set of decision rules. For example, decision trees have been adopted to explain neural networks \cite{krishnan1999extracting} and tree ensembles \cite{hara2016making,tan2016tree}, while decision rules have been widely used to explain neural networks \cite{augasta2012reverse,andrews1995survey} and support vector machines \cite{fung2005rule}. A few methods for global explanation are agnostic w.r.t.\ the learning model \cite{lou2012intelligible,henelius2014peek}.

A different stream of approaches, still in the XbD setting, focuses on the \textit{local} behavior of a black box \cite{guidotti2018survey}, searching for an explanation of the decision made for a specific instance. Some such approaches are model-dependent and aim, e.g., at explaining the decisions of neural networks by means of saliency masks, i.e., the portions of the input record (such as the regions of an input image) that are mainly responsible for the classification outcome \cite{xu2015show,zhou2016learning}. 
A few more recent methods are model-agnostic, such as LIME \cite{ribeiro2016should}. The main idea is to derive a local explanation for a decision outcome $y$ on a specific instance $x$ by learning an interpretable model from a randomly generated neighborhood of $x$, where each instance in the neighborhood is labelled by querying the black box. An extension of LIME using decision rules (called Anchors) is presented in \cite{ribeiro2018anchors}, which uses a bandit algorithm that randomly constructs the rules with the highest coverage and precision.

When the training set is available, decision rules are also widely used to proxy a black box model by directly designing a transparent classifier \cite{guidotti2018survey} which is locally or globally interpretable on its own \cite{lakkaraju2016interpretable,malioutov2017learning}.

To sum up, despite the soaring attention to the topic, the state of the art to date still exhibits ad-hoc, scattered results, mostly hard-wired with specific models. A widely applicable, systematic approach with a real impact has not emerged yet. In our view, a black box explanation framework should be: 
\begin{enumerate}
    \item \emph{model-agnostic}, so it can be applied to any black box model;
    \item \emph{logic-based}, so that explanations can be made comprehensible to humans with diverse expertise, and support their reasoning;
    \item \emph{both local and global}, so it can explain both individual cases and the overall logic of the black-box model;
    \item \emph{high-fidelity}, so it provides a reliable and accurate approximation of the black box behavior.
\end{enumerate}
The four desiderata do not coexist in current proposals.
Logic-based decision rules have proven useful in the sub-problem of explaining discrimination from a purely data-driven perspective, as demonstrated in the lively stream of research in discrimination-aware data mining, started in \cite{pedreshi2008discrimination,RuggieriPT10}, but it is unlikely that rules in their simplest form will solve the general explanation problem. Global rule-based models, trained on black box decision records, are often either inaccurate, over-simplistic proxies of the black box, or too complex, thus compromising interpretability. On the other hand, purely local models, such as LIME, do not yield an overall proxy of the black box, hence cannot solve the XbD and BBX problems in general terms.

Here we propose to tackle the problem from a different perspective: more expressive rule languages equipped with novel rule learning methods, realizing a different, local-first mix of the local and global methods. This is the focus of the rest of the paper.

\section{How to construct meaningful explanations?}
\label{sec:toe}


Let us consider the XbD problem of discovering an explanation for a high-quality, non overfitting black box model $b$ learned over a training dataset $D$ of labelled examples $(x,y)$, where $y$ is the class label and $x$ is a vector of observed features; let us concentrate on binary classification, i.e., $y \in \{0,1\}$. Our framework works under three assumptions.

\begin{itemize}
    \item {\b H1: Logic explanations.} The cognitive vehicle for offering explanations should be as close as possible to the language of reasoning, that is \emph{logic}. From simple propositional rules up to more expressive, possibly causal, logic rules, many options of varying expressiveness exist to explore the trade-off between accuracy and interpretability of explanations.
    \item {\b H2: Local explanations.} The decision boundary for the black box $b$ can be arbitrarily complex over the whole training dataset $D$, but \emph{in the neighborhood of each specific data point $(x,y) \in D$ there is a high chance that the decision boundary is clear and simple}, hence amenable to be captured by an interpretable explanation. 
    \item {\b H3: Explanation composition.} There is a high chance that \emph{similar data points admit similar explanations}. Also, similar explanations are amenable to be composed together to yield global explanation of the black box.
\end{itemize}

H2 is motivated by the observation that if all data points in the training set are surrounded by complex decision boundaries, then the black box $b$ is likely to be in overfitting, unable to generalize from a training dataset of insufficient quality, thus contradicting the basic assumption. Analogous contradiction holds for H3, if any two data points admit very different explanations due to different decision boundaries. These assumptions suggest a two-step, \emph{local-first} approach to the XbD explanation problem:

\begin{enumerate}
    \item \textit{(local step)} For any example in the training dataset $D$ relabeled by the black box $b$, i.e.\ for any specific $(x,y')$, where $y'=b(x)$ is the label assigned by $b$ to $x$: query $b$ to label a sufficient set of examples (\textit{local dataset}) in the neighborhood of $(x,y')$ which are then used to derive an explanation $e(x, y')$ for $b(x) = y'$. The explanation answers the question: why $b$ has assigned class $y'$ to $x$? and, possibly, also its counterfactual: what should be changed in $x$ to obtain a different classification outcome?
    \item \textit{(composition step)} Consider as an initial global explanation the set of all local explanations $e(x,y')$ constructed at the local step for each individual example $(x,y')$ and synthesize a smaller set by iteratively composing and generalizing together similar explanations.
\end{enumerate}

We discuss next the key issues of ($i$) a language for explanations, ($ii$) the inference of local explanations, and ($iii$) the bottom-up synthesis of global explanations.

\subsection{Rules languages for explanation}

An explanation is a comprehensible representation of a decision model associated with a black box, acting as an interface between the model and the human. According to assumption H1, an explanation can be specified through decision rules expressed in various logic-based languages, characterized by (i) the form of predicates and constraints over features admissible in the rules, and (ii)  statistical measures of confidence associated with the rules. 

As noted in Section 2, simplistic decision rules may not be expressive enough for proxying the decision behaviour of sophisticated machine learning models, such as deep neural networks, support vector machines or ensemble methods. Rule languages for explanation should range from simpler to more expressive alternatives: 
\begin{itemize}
    \item plain association rules on nominal features, e.g., \texttt{\small CheckingBalance=low, SavingBalance=low $\rightarrow$ Credit=no},
    \item decision rules on features of generic type, e.g., \texttt{\small CheckingBalance $<$ 0, Sa\-vingBalance $<$ 100 $\rightarrow$ Credit=no},
    \item rules with inter-feature constraints, with reference to constraint languages of increasing complexity, e.g., \texttt{\small CheckingBalance $>$ 0, SavingBalance $>$ 100, CreditBalance $+$ SavingBalance $< 200$ $\rightarrow$ Credit=no},
    \item rules with parameter features, which abstract a collection of inter-feature constraints, e.g., \texttt{\small CreditBalance $+$ SavingBalance $\geq$ a, CreditBalance $-$ SavingBalance $\geq 500-$a, 200 $\geq$ a $\geq$ 300 $\rightarrow$ Credit=no}. 
\end{itemize}

Explanations might also be equipped with \emph{counterfactuals} \cite{wachter2017counterfactual}, sets of rules with opposite decision and minimal change of the premise of a specific  decision rule, in order to characterize under which slightly different conditions the conclusion of a rule is reverted. Examples of explanations with counterfactual are reported later. Finally, the above list could be extended by resorting to more expressive logics to deal with causation and/or time. The Probabilistic Computation Tree Logic \cite{kleinberg2009temporal}, for instance, allows expressing Suppes probabilistic causation. Approaches that infer formula from temporal data have been successfully applied in the context of understanding cancer progression \cite{caravagna2016algorithmic}. However, a critical point is to balance the expressiveness of the logic used with the computational complexity of reasoning over formulae in the logic. Entailment is decidable in polynomial time for linear constraints, but it becomes co-NP hard for parameterized linear constraints (see \cite{eirinakis2012complexity} for negative result and tractable fragments). More generally, a novel avenue of research opens here: causal inference and learning (see \cite{peters2017elements}) applied to the outcome of queries to a black box aimed at revealing the hidden causal structure implied by the black box when applied in the ``real world". Causal explanations highlight which conditions on a sample $x$ actually determine the black-box decision $b(x)$. This is a central problem when correlation is not enough, e.g.,~in the context of discrimination litigation~\cite{Foster2004}.

Statistical measures associated to a rule $A \rightarrow B$ also range from simple to complex:
\begin{itemize}
	\item Rule support $P(A,B)$, coverage $P(A)$; and confidence, i.e., the conditional probability $P(B \mid A)=\frac{P(A,B)}{P(A)}$;
	\item Rule lift: $\frac{P(A,B)}{P(A)P(B)}$ and other correlation scores, such as the maximum mutual information, the reduction in uncertainty of $B$ when $A$ is known: $S(A,B)= \frac{H(A)+H(B)-H(A,B)}{min[H(A),H(B)]}$ where $H$ denotes the entropy (or uncertainty) of a variable;
	\item Statistical tests of the significance of the previous measures, w.r.t.\ various choices of null models/hypotheses \cite{Fleiss2003}.
    \item Causal extensions of the previous correlation-based measures, e.g., by  \emph{propensity score} reweighing \cite{qureshi2016causal} or by probabilistic causation confidence score derived from Suppes-Bayes causal networks \cite{bonchi2017exposing}.
\end{itemize}

Operators which manipulate explanations together with their associated measures are also needed, such as composition operators that  merge rules, in order to synthesize global explanations from local ones as we are explaining in the next section, as well as generalization operators that lift a collection of rules to a higher level of abstraction. An example of generalization consists in introducing parameters. E.g.,~the rules \texttt{\small CreditBalance $\leq$ 200, CheckingBalance $\leq$ 300 $\rightarrow$ Credit=no} and \texttt{\small CreditBalance $\leq$ 300, CheckingBalance $\leq$ 200 $\rightarrow$ Credit=no} have the minimal affine generalization \texttt{\small CreditBalance $\leq a$, CheckingBalance $\leq$ 500$-a$, 200 $\leq a \leq$ 300 $\rightarrow$ Credit=no}. In such an example, the learned parameter $a$ may reveal a latent feature used in decision-making. Details of the approach for learning parameterized linear systems are reported in our previous work \cite{ruggieri2013learning}.








\subsection{Local explanations} 
The \textit{local-first} approach that we propose requires the extraction of local explanations to be merged with some mechanism in order to get a global explanation. In the literature, as discussed above, there exist some approaches for finding local explanations like those presented in \cite{guidotti2018local,ribeiro2016should,ribeiro2018anchors}. They aim at returning an individual explanation $e$ for the decision assigned to each record $x \in X$ by the black box.
Given a record $x$ to be explained, those explanators return a local explanation $e(x,y')$ (where $y' = b(x)$) 
reasoning on a local dataset $N(x,y')$, generated in the neighborhood of $(x,y')$ using the black box $b$ to assign class labels to the instances in the neighborhood. In particular, on top of $N(x,y')$, they build an interpretable classifier $c(x) = y'$ from which it is possible to derive the explanation $e(x, y')$.
These approaches mainly differ from each others on both: \emph{(i)} the procedure used to create the local training dataset $N(x,y')$, and \emph{(ii)} the derived interpretable classifier $c$.

\begin{figure}[t]
    \centering
    \hspace{-2mm}
   		\includegraphics[trim = 2mm 0mm 1mm 0mm, clip,width=0.4\linewidth]{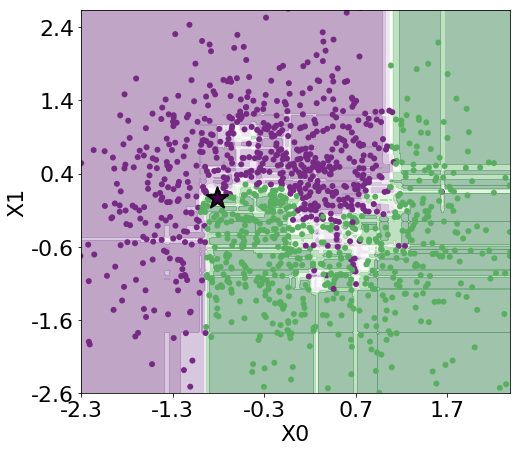}\hspace{2ex}
        \includegraphics[trim = 2mm 0mm 1mm 0mm, clip,width=0.4\linewidth]{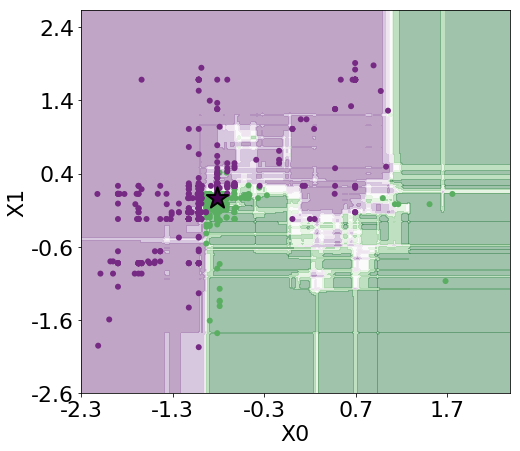}\\
        \includegraphics[trim = 2mm 0mm 1mm 0mm, clip,width=0.4\linewidth]{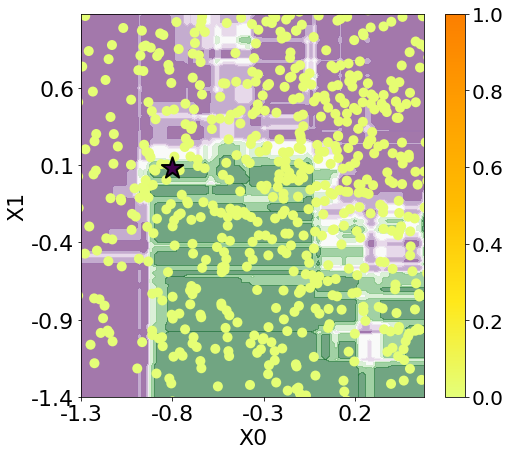}\hspace{2ex}
        \includegraphics[trim = 2mm 0mm 1mm 0mm, clip,width=0.4\linewidth]{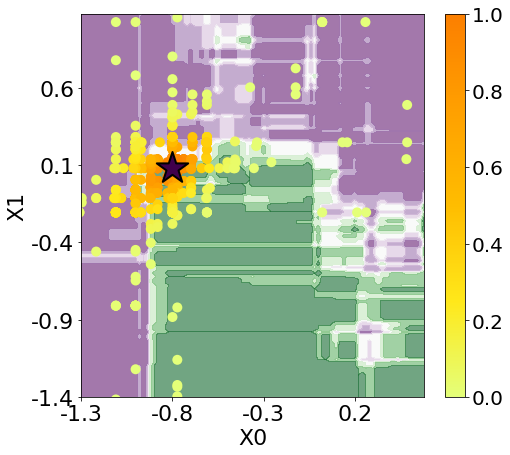} 
\caption{
From \cite{guidotti2018local} Black box decision purple vs green boundary. Starred instance to explain $x$. 
Uniformly random (left) and genetic generation (right).
The bottom figures show the density scatter plot of a zoomed area around $x$ of the top figures.
(Best view in color).}
\label{fig:neigh}
\vspace{-2mm}
\end{figure}

In particular, \textit{LIME}~\cite{ribeiro2018anchors} uses a purely random neighborhood generation (see Figure \ref{fig:neigh} (left)) and as interpretable classifier $c$ a linear model, while the weights of the coefficients (i.e., the features importance) form the explanation $e$ (see \cite{guidotti2018survey} for more details).
\textit{Anchor}~\cite{ribeiro2018anchors} uses a bandit algorithm for the neighborhood generation that randomly constructs the anchors.
An anchor is a decision rule, i.e.,~the explanation $e$, that sufficiently ties a prediction locally such that changes to the values of features not in the rule do not affect the decision outcome.
In~\cite{guidotti2018local} we propose \textit{LORE (LOcal Rule-based Explanations)}.
LORE uses a genetic algorithm approach to generate the neighborhood $N(x,y')$ (see Figure \ref{fig:neigh} (right)) and a decision tree as interpretable classifier $c$, while the explanation consists in a rule derived from the decision tree classifier by following the path from the root to a leaf according to the values of $x$.
Moreover, LORE also returns a set of counterfactual rules, suggesting the changes in the instance's features of $x$ that may lead to a different outcome.

\begin{figure}[t]
\vspace{2mm}
{\bf - LIME} \hfill
\vspace{-2mm}
\begin{minipage}[c]{\linewidth}
\centering
\includegraphics[trim = 0mm 0mm 0mm 0mm, clip,width=0.5\linewidth]{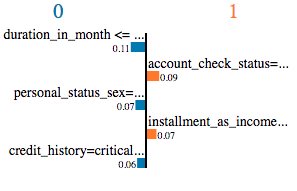}
\end{minipage}\\
{\bf - Anchor} \hfill
\vspace{3mm}
\begin{minipage}[c]{\linewidth}\em
\begin{tabular}{lp{60ex}}
a = & (\{credit\_history = critical account,\\ & duration\_in\_month $\in$ [0, 18.00]\}
       $\rightarrow$ decision = 0)
\end{tabular}
\end{minipage}
\vspace{2mm}
{\bf - LORE} \hfill
\vspace{1mm}
\begin{minipage}[c]{\linewidth}\em
\begin{tabular}{lp{60ex}}
r = & (\{credit\_amount $\leq$ 836,
      housing = own, other\_debtors =\\ & \quad none,
      credit\_history = critical account\}
       $\rightarrow$ decision = 0)\\  
$\Phi$ = & \{ (\{credit\_amount $>$ 836, 
      housing = own, other\_debtors =\\ & \quad none, 
      credit\_history = critical account\}
     $\rightarrow$ decision = 1),\\
  &   (\{credit\_amount $\leq$ 836,
       housing = own, other\_debtors =\\ & \quad none, credit\_history = all paid back\}
      $\rightarrow$ decision = 1) \}\\
\end{tabular}
\end{minipage}\\
\caption{Example of explanations of LORE, LIME and Anchor from \cite{guidotti2018local}.}
\label{fig:expl_example}
\vspace{-3mm}
\end{figure}

To better understand these different ways of providing explanations for a record $x$ we report in Figure~\ref{fig:expl_example} the three local explanations for an instance $x$ of the well-known \texttt{german} dataset \cite{german_dataset} from UCI\footnote{\url{https://archive.ics.uci.edu/ml/index.php}}. 
The top explanation is by LIME. 
Weights are associated to the categorical values in the instance $x$, and to continuous upper/lower bounds where the bounding values are taken from $x$. 
Each weight tells the user how much the decision would have changed for different (resp., smaller/greater) values of a specific categorical (resp., continuous) feature. 
LIME explanations are not straightforward to follow, compared to rule-based explanations of Anchor and LORE.
The central explanation in Figure~\ref{fig:expl_example} is by Anchor, a single decision rule characterizing the contextual conditions for the decision of the black box. Anchor requires a discretization of continuous features,
The bottom explanation in Figure~\ref{fig:expl_example} is by LORE.
Rule $r$ inherits the expressiveness of decision tree split conditions, e.g.,~on continuous features. 
Moreover, the counterfactual rules $\Phi$ provides high-level and minimal-change contexts for reversing the outcome prediction of the black box.


\subsection{From local to global explanations}
\label{sec:dendrogram}

Instead of learning directly a global interpretable model that tries to imitate the black box, an alternative, more promising approach is to synthesize a global explanation from the bottom-up, starting from the collection of all local explanations, as discussed above. While many realizations of this idea are possible, here we discuss a natural one: the bottom-up construction of a \emph{dendrogram}, a binary tree describing the compositions of pairs of (similar) explanations into a single, more general explanation, to be used as a means to find, approximately, an optimal collection of explanations to proxy the overall behavior of a black box.

Two basic functions over explanations are needed: a \emph{distance} function $d(e,e') \in [0,1]$ (with $d(e,e')=0$ meaning that $e$ and $e'$ are identical, and $d(e,e')=1$ meaning that $e$ and $e'$ are disjoint; and a \emph{merge} function $m(e,e')=e''$ mapping $e$ and $e'$ into a (minimal) generalized explanation $e''$ that subsumes both. A bottom-up algorithm to construct a dendrogram starting from the collection $E = \{e(x,y) \mid (x,y) \in D\}$, i.e., all local explanations, is the following.

\begin{enumerate}
    \item set $E = \{e(x,y) \mid (x,y) \in D\}$
    \item select $e$ and $e'$ such as $d(e,e')$ is minimal in $E$, i.e., the two most similar explanations in $E$ according to $d$
    \item set $e'' = m(e,e')$, i.e., merge $e$ and $e'$,
    \item set $E = E \setminus \{e,e'\} \cup \{e''\}$, i.e., replace $e$ and $e'$ in $E$ with $e''$,
    \item add a merge node in the dendrogram between the nodes corresponding to $e$ and $e'$ at height $d(e,e')$,
    \item repeat steps 2--5 until $E$ contains a single explanation.
\end{enumerate}
	
\begin{figure}[!t]
  \centering
    \includegraphics[width=0.7\textwidth]{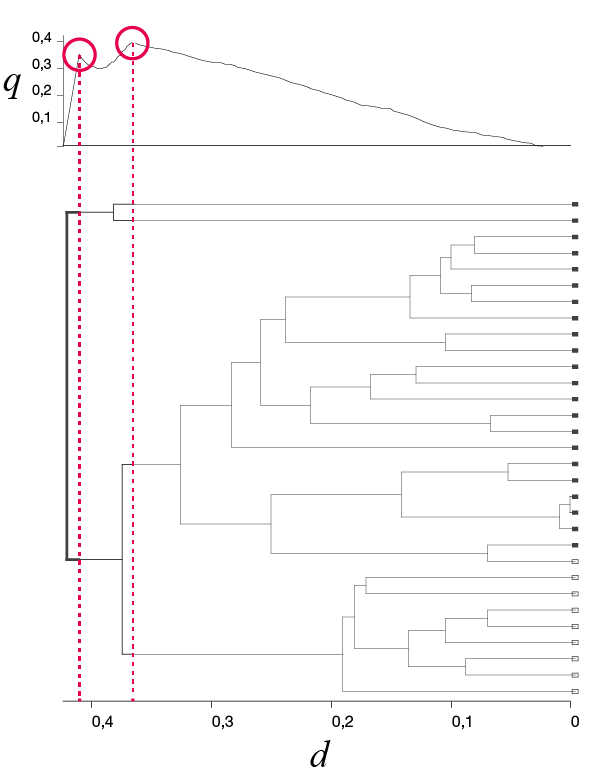}
    \caption{\textbf{Bottom-up synthesis of global explanations.}}
    \label{fig:dendrogram}
\end{figure}

The final task is now to exploit the dendrogram to find an optimal global explanation, that is a collection $\hat{E}$ of explanations extracted from the dendrogram that  covers all initial local explanations and maximizes an appropriate quality score $q(E)$ that, for any collection $E$ of explanations, measures both (i) the fidelity achieved adopting $E$ for mimicking the black box $b$, and (ii) the number and size of the explanations in $E$, i.e., the complexity of the collection. Clearly, the goal is to maximize fidelity while minimizing complexity. Many alternatives are conceivable to define $q$, such as variants of the Minimum Description Length criterion (MDL), or the Bayesian Information Criterion (BIC). Also, alternative cutting or pruning methods can be used to identify the best collection of explanations in the dendrogram. Figure \ref{fig:dendrogram} illustrates a possibility:
compute the value of $q(E)$ for all $E$ resulting by cutting the dendrogram at all splitting points, and select the cut whose corresponding collection maximizes $q$.

As a preliminary experimental validation of the local-first approach to global explanations, we compared it with two baselines: pure global classification \cite{craven1996extracting,domingos1998knowledge}, i.e., learning a decision tree on the whole training dataset, and the collection of all local explanations obtained applying the local \emph{explanators} LORE and Anchor \cite{guidotti2018local,ribeiro2018anchors} separately to each example in the training dataset (removing duplicate rules). Each method works on the set of instances in the training, relabelled by the decisions assigned by the black box $b$ (a random forest predictor in the experiment). To realize the dendrogram bottom-up method we start with the local rules generated by LORE and Anchor and apply a merge operator inspired to that of \cite{Ruggieri13}. The distance function between explanations, needed to construct the dendrogram, is based on the Jaccard distance, computed on the two sets of data records covered by the explanations. We use BIC to identify the optimal global explanation, measuring the explanation complexity by the number of rules. For rule classification we use the CPAR strategy of weighted voting proposed in \cite{yin2003cpar}. 



We discuss, for instance, the results on the application of the three methods to the UCI dataset \texttt{compas} containing the features used by the COMPAS algorithm for scoring the crime recidivism risk of defendants for more than 10,000 individuals. We considered the binary problem of classifying ``Low-Medium'' and ``High'' risk. Although the adopted design choices for the local-to-global method are first-cut and a many unexplored alternatives exist, we observe that our local-to-global method produces a model with a comparable fidelity to the other two but, especially using LORE, with one order of magnitude less rules (from $\approx 600$ to $\approx 50$ rules). This is a remarkably promising  outcome, that calls for far more extensive empirical validation along a wide number of design options.

%

\section{Explanation Discovery and Reasoning: Research Directions}
\label{sec:alg}

The proposed framework of local-first, bottom-up discovery of explanations based on highly expressive rules has a large potential for future research: a virtuous cycle may start, where more expressive, high-level rule languages call for novel ways to query and audit black box models and novel algorithms for learning rule in statistical and causal terms, which in turn may suggest novel ways to generalize rules and to devise reasoning mechanisms on top of. Higher-level rules will require new rule discovery algorithms, way beyond the state of art of association rule based classification algorithms \cite{yin2003cpar}, or traditional rule learning algorithms \cite{FurnkranzK15}. Moreover, the approach needs to be extended also beyond (binary) classification, i.e., to ordinal classification and for predicting numeric variables (e.g., regression rules), in order to deal also with ranking and scoring problems. 

Regarding reasoning mechanisms, simply providing the final user with the set of explanations computed by the discovery algorithms may not suffice, also depending on the expertise of the user, or on the need to interact with the explanations by asking high-level questions. Example questions include: For what reasons was my application rejected? What are the rules that apply to a specific population or profile? Which rules hide potential discrimination related to, e.g., protected minorities? How do the confidence or other statistical and causal properties of a rule vary by changing a threshold value in the rule antecedent, or by dropping or adding a constraint? What combinations of features are most strongly correlated to (or are a cause of) a specific decision outcome? This final example question might be aimed, e.g., at discovering novel forms of discrimination towards vulnerable groups or profiles, or highlighting spurious rules due to artifacts in the collected training data. Based on the the algebraic properties of rule languages, the design of reasoning mechanisms for rule manipulation and filtering can provide meaningful answers for the final users. Appropriate interfaces must also be provided, including visual and textual presentation, as well as visual exploration for online analytics. 

\subsection*{Text, images and non-relational data} 
The discussion so far has focused on relational data, characterized by meaningful features. Current systems, however, deal with a heterogeneous variety of data sources, such as networks, spatio-temporal trajectories, text, images and multimedia. In fact, black box models such as deep learning and complex neural networks have shown a notable classification power in, e.g., image recognition problems. How can the explanation discovery process described in this paper be generalized to other, poorly structured, forms of data? An interesting aspect to study is how far the agnostic approach can be pushed by exploiting the results in semantic annotation that are rapidly emerging in the related fields, in order to map raw image data, text data, etc. into collections of meaningful objects. The idea is then to use this semantic transformation of decision records as an explanation set in the discovery process, in order to produce comprehensible rules. One example of a semantic annotation tool in the text mining domain is TagMe \cite{FerraginaS12}, which maps selected keywords in a text to Wikipedia concepts. Similar tools capable of mapping image parts to meaningful concepts in a systematic way, when available, have the potential of generalizing the ad-hoc strategies adopted, e.g., in \cite{ribeiro2016should} and automatically discovering biased rules such as “if there is a big white zone in the picture behind the animal, then it is a wolf”. The adoption of semantic annotation tools can be used a pre-processing module for the explanation discovery process. One may object that, following our approach, the XbD or BBX problem is tackled relying on semantic annotation black boxes, thus explaining a black box through other black boxes. We observe that, in the end, explanation is always a translation of a complex object in terms of simpler ones, that the user can understand or, at least, fully trust. Therefore, explaining a black-box through simpler ones that are already understood or trusted is a viable approach in tasks such as object recognition, natural language understanding, etc., as higher-quality, fully validated machine learning basic components will become available.


\subsection*{Hidden features and background information} 
In the BBX problem, a very challenging aspect is that the black box may use more information than explicitly asked the user, e.g., by inferring further features from the user’s input and other available sources. Also, the decisions of a black box might be better understood if the indirect inferences adopted by the system, either consciously or not, are made explicit and brought to light. One example is indirect discrimination, as in the Amazon.com case, inadvertently using redlining rules that prevented minority neighbourhoods from participating in a program offering free same-day delivery. An interesting aspect to be investigated is to understand how the information on the explanation set can be extended with supplementary features from the wealth of open data available from official statistics and demographic institutes and other public organizations. The rule composition operators will provide means to merge background information and original features into learned statistical rules, e.g., expanding the ideas of the framework for inferring indirect discrimination rules described in \cite{RuggieriPT10}. Following previous approaches, the rule transformation operators of the algebra will enable, for instance, multiple rules such as \texttt{\small ZIP=c $\rightarrow$ FreeDelivery=no} combined with background information \texttt{\small ZIP=c $\rightarrow$ MinorityNeighborhood=yes} to be mapped into a new general rule \texttt{\small MinorityNeighborhood=yes $\rightarrow$ FreeDelivery=no} together with bounds on its statistical confidence and causal validity.



\section{Conclusions}
\label{sec:conclusion}

The local-to-global framework for black box explanation, introduced in this paper, paves the road for a wide spectrum of further research works along three dimensions. First, the language for expressing explanations, which may draw inspiration from the rule-based, declarative languages developed since the 80's, such as constraint and inductive logic programming, as well as from the ideas of causal logic. Second, the inference of local explanations aimed at a specific decision instance, which calls for exploring alternative ways to query and audit the black box in the vicinity of the target instance to the purpose of revealing the statistical and causal logic adopted; this can possibly draw from the body of research in active learning and testing, such as fairness testing \cite{GalhotraBM17}. Third, the bottom-up generalization of the local explanations into global ones, which calls for algorithms that optimize the quality of explanations in terms of fidelity, simplicity, and coverage. 

A few recent proposals, including ours, are initial seeds along this road to design a systematic, agnostic method, i.e., one that does not take into account the internals of the decision model, even if they are known, as in the XbD problem. This widens applicability to real cases of the BBX problem, allowing dealing with generic decision models, which do not necessarily involve machine learning, but generally algorithms, humans or a mixture of them. In the XbD problem, the approach allows a data scientist to use, in principle, any kind of machine learning model. 

A critical aspect for this research endeavor in the general BBX problem is that it requires the availability of decision record data, i.e., examples to fuel the explanation discovery process. An interesting complementary research activity is how to favor the collection of data through participatory \emph{watchdog platforms}, enabling users or consumers subject to automated decision-making to share their own decision records within a privacy-preserving, crowd-sourcing framework, in order to accumulate sufficient evidence for the explanation discovery process and expose the profiling logic of the black box. This would expand the applicability of the explanation technology beyond the XbD case, potentially helping to re-balance the information asymmetry between individual users and ``big data'' companies.

In fact, a technology for the explanation of black boxes would have a strong ethical impact. It may empower individuals against undesired, possibly illegal, effects of automated decision-making systems which may harm them, exploit their vulnerabilities, and violate their rights and freedom. It may provide practical tools for implementing the “right of explanation” provisions of the European GDPR, provided it delivers intuitive and usable explanations to users with different levels of expertise. It may improve industrial procedures and standards for the development of  services and products powered by machine learning components, thus increasing the trust of companies and consumers in AI-powered products. It may empower citizens and policy makers with the ability of discovering new forms of discrimination towards vulnerable social groups and improving anti-discrimination norms and practice.

We are evolving, faster than expected, from a time when humans are coding algorithms and carry responsibility of the resulting software quality and correctness, to a time when machines automatically learn algorithms from sufficiently many examples of the algorithms' expected input/output behavior. Requiring that machine learning and AI be explainable and comprehensible in human terms is not only instrumental for validating quality and correctness, but also for aligning the algorithms with human values and expectations, as well as preserving human autonomy and awareness in decision making.

\section{Acknowledgement}
This work is partially supported by the European Community's H2020 Program under the funding scheme ``INFRAIA-1-2014-2015: Research Infrastructures'' grant agreement 654024, \url{http://www.sobigdata.eu}, ``SoBigData: Social Mining \& Big Data Ecosystem''.

\bibliographystyle{ACM-Reference-Format} 
\bibliography{biblio}

\end{document}